\documentclass[conference]{IEEEtran}
\IEEEoverridecommandlockouts
\usepackage{cite}
\usepackage{amsmath,amssymb,amsfonts}
\usepackage{graphicx}
\usepackage{textcomp}
\usepackage{xcolor}
\usepackage{algpseudocode}
\usepackage{algorithm}
\def\BibTeX{{\rm B\kern-.05em{\sc i\kern-.025em b}\kern-.08em
    T\kern-.1667em\lower.7ex\hbox{E}\kern-.125emX}}
\usepackage{multirow}
\makeatletter 
\newcommand{\linebreakand}{%
  \end{@IEEEauthorhalign}
  \hfill\mbox{}\par
  \mbox{}\hfill\begin{@IEEEauthorhalign}
}
\makeatother 

\usepackage[skip=2pt]{caption}

\usepackage{booktabs}
\usepackage{gensymb}
\usepackage{geometry}
\begin{document}

\title{A Comparison of Classical and Deep Reinforcement Learning Methods for HVAC Control\\
{}
\thanks{}
}

\author{\IEEEauthorblockN{Marshall Wang}
\IEEEauthorblockA{\textit{Vector Institute} \\
Toronto, Canada \\
marshall.wang@vectorinstitute.ai}
\and
\IEEEauthorblockN{John Willes}
\IEEEauthorblockA{\textit{Vector Institute} \\
Toronto, Canada \\
john.willes@vectorinstitute.ai}
\and
\IEEEauthorblockN{Thomas Jiralerspong
}
\IEEEauthorblockA{\textit{Vector Institute} \\
Toronto, Canada \\
thomasjiralerspong@gmail.com}
\and
\linebreakand
\IEEEauthorblockN{Matin Moezzi}
\IEEEauthorblockA{\textit{Vector Institute} \\
Toronto, Canada \\
matin.moezzi@mail.utoronto.ca }
}

\maketitle

\begin{abstract}
Reinforcement learning (RL) is a promising approach for optimizing HVAC control. RL offers a framework for improving system performance, reducing energy consumption, and enhancing cost efficiency. We benchmark two popular classical and deep RL methods (Q-Learning and Deep-Q-Networks) across multiple HVAC environments and explore the practical consideration of model hyper-parameter selection and reward tuning. The findings provide insight for configuring RL agents in HVAC systems, promoting energy-efficient and cost-effective operation.
\end{abstract}


\section{Introduction}

Heating, ventilation, and air conditioning (HVAC) play a fundamental role in our infrastructure. They are essential processes that come with substantial environmental and economic costs. Whether it's ensuring optimal temperature control in datacenters or maintaining comfortable indoor environments, the accurate management of thermodynamic system properties is crucial. However, there is an opportunity to develop dynamic HVAC control systems that minimize energy consumption and improve cost efficiency while meeting operational constraints.

Reinforcement learning (RL) has emerged as a promising approach for HVAC control \cite{mason2019review, sierla2022review}. By formulating HVAC control as a sequential decision-making problem with a well-defined reward structure, RL provides a powerful framework to optimize system performance. RL can be employed to directly control HVAC system components or utilized to control setpoints of an inner-loop controller. Direct control involves training an RL agent to autonomously adjust parameters such as fan speeds and damper positions, optimizing their settings based on the observed state of the system and desired outcomes. In a setpoint control configuration, the RL agent serves as an intelligent supervisor, influencing the heating and cooling setpoints provided to a traditional controller that operates the HVAC system. 

Despite the promising results \cite{faddel2020data, wei2017deep}, challenges remain in implementing RL-based HVAC control. Training RL agents in realistic and dynamic environments can be challenging for machine learning practitioners \cite{sierla2022review}. The need for accurate system modeling, data availability, and considerations for safety and stability poses further complexities in the practical deployment of RL solutions for HVAC control. This work seeks to provide a practical exploration of popular classical and deep reinforcement learning methods across multiple HVAC environments. The core contributions of this work are: 

\begin{enumerate}
    \item Benchmarking of the popular Q-Learning \cite{watkins1992q} and Deep-Q-Network (DQN) \cite{mnih2013playing} algorithms for HVAC control across multiple building environments and geographic locations.
    \item A practical exploration of model hyper-parameter and reward function tuning to guide machine learning practitioners when configuring reinforcement learning agents in their environments.
\end{enumerate}


This work is structured as follows: Section \ref{sec:related-work} provides an overview of HVAC control, building simulation, and reinforcement learning for HVAC control. Section \ref{sec:methods} provides an overview of Q-Learning and DQN methods. In Sections \ref{sec:experiments} and \ref{sec:results}, we investigate the relative performance and configuration of these methods in a set of simulated building and geographic environments.

\section{Related Work} \label{sec:related-work}

\subsection{Reinforcement Learning for HVAC Control}

Reinforcement learning has demonstrated utility across many domains including robotics \cite{kober2013reinforcement, singh2022reinforcement}, large language models \cite{ouyang2022training}, and gaming \cite{silver2016mastering}. RL also offers an effective framework for optimizing HVAC system performance by formulating control as a sequential decision-making problem, where RL agents can either directly control system components or influence setpoints provided to a traditional controller \cite{sierla2022review}. 

One of the most popular reinforcement learning algorithms, Q-Learning, has been extensively applied to HVAC control \cite{cho2006application, faddel2020data, li2015multi, barrett2015autonomous}. \cite{li2015multi} propose a multi-grid Q-Learning method to mitigate slow convergence and by first adopting a coarse model to fast converge then adopting a fine model to further improve optimization. \cite{lissa2020transfer} explore the impact of spatial changes on the performance of Q-Learning HVAC control in buildings and proposes an adaption via transfer learning, resulting in significantly reduced learning time and improved control policies, with minimal impact on user comfort, particularly in locations with lower temperature variation. 

Following the introduction of Deep-Q-Networks \cite{mnih2013playing} in 2013, they were quickly adapted for use in HVAC system control. \cite{wei2017deep} were one of the first to extend them to an HVAC control domain. \cite{jia2019advanced} learn an end-to-end deep reinforcement learning approach which regulates the smoothness of the neural network to penalize erratic behavior, resulting in a stabilized learning process and the development of interpretable control laws. \cite{yu2020multi} propose an HVAC control algorithm which formulates energy minimization in complex multi-zone systems as a Markov game. The Markov game is then solved using multi-agent deep reinforcement learning with an attention mechanism.


\subsection{Building Energy Simulation}

Building Energy Simulation (BES) is an important task to support the design of energy efficient infrastructure, however, they also provide a testbed for the development of intelligent agents for dynamic HVAC control. BES software allows for fine-grained configuration of building layout and HVAC systems. It encompass a range of tools, including EnergyPlus, TRNSYS, CarnotUIBK among many others \cite{magni2021detailed}. EnergyPlus \cite{crawley2001energyplus} is one of the most widely adopted by architectural design software packages such as OpenStudio and AutoDesk Revit. 

Frameworks such as Sinergym \cite{2021sinergym} and Energym \cite{scharnhorst2021energym} and RL Testbed \cite{moriyama2018reinforcement} have also been developed using EnergyPlus as the underlying simulation engine, providing a Python API for easy integration with open-source reinforcement learning packages such as RLib and StableBaselines. These frameworks integrate with EnergyPlus models, allowing for the training of reinforcement learning agents, reproducibility of experiments, and benchmarking of building control methods across diverse environments and action spaces. 

\begin{algorithm}[t!]
  \caption{Q-learning algorithm}
  \label{alg:q-learning}
  \begin{algorithmic}[1]
    \State Initialize $Q(s,a)$ arbitrarily
    \State Set learning rate $\alpha \in (0,1]$
    \State Set discount factor $\gamma \in (0,1]$
    \State Set exploration rate $\epsilon \in (0,1]$
    \For{each episode}
        \State Initialize $s$
        \While{$s$ is not terminal}
            \If{random $< \epsilon$}
                \State $a \gets$ random action
            \Else
                \State $a \gets \arg\max\limits_{a'} Q(s,a')$
            \EndIf
            \State Take action $a$, observe $r$, $s'$
            \State $Q(s,a) \gets (1 - \alpha) Q(s,a) + \alpha [r + \gamma \max\limits_{a'} Q(s',a')]$
            \State $s \gets s'$
        \EndWhile
    \EndFor
  \end{algorithmic}
\end{algorithm}
\begin{algorithm}[t!]
\caption{Deep Q-Network (DQN)}
\label{alg:dqn}
\begin{algorithmic}[1]
\State Initialize replay memory $D$ to capacity $N$
\State Initialize action-value function $Q$ with random weights $\theta$
\State Initialize target action-value function $\hat{Q}$ with weights $\theta^{-} = \theta$
\For{$episode=1$ to $M$}
\State Initialize state $s_1$
\For{$t=1$ to $T$}
\State With probability $\epsilon$ select a random action $a_t$
\State otherwise select $a_t=\arg\max_{a}Q(s_t,a;\theta)$
\State Execute action $a_t$ in environment and observe reward $r_t$ and next state $s_{t+1}$
\State Store transition $(s_t,a_t,r_t,s_{t+1})$ in $D$
\State Sample a minibatch of transitions $(s_i,a_i,r_i,s_{i+1})$ from $D$
\State Set target $y_i=r_i+\gamma \max_{a'}\hat{Q}(s_{i+1},a';\theta^{-})$
\State Update weights: $\theta \leftarrow \theta - \alpha \nabla_{\theta} \left(Q(s_i,a_i;\theta)-y_i\right)^2$
\State Every $C$ steps reset $\hat{Q}=Q$
\EndFor
\EndFor
\end{algorithmic}
\end{algorithm}

\section{Methods} \label{sec:methods}

We will now define the HVAC learning problem and present the Q-Learning and Deep-Q-Networks methods in depth.

\subsection{Problem Formulation}

Reinforcement learning is a machine learning paradigm where an agent learns to take actions by interacting with the environment and receiving positive or negative feedback \cite{sutton2018reinforcement}. The goal for the agent is to learn to maximize cumulative rewards over time through appropriate action selection based on observed state and received rewards. Reinforcement learning comprises of several basic elements: agent, environment, state, action, reward, policy. The agent is the learning entity making decisions, and the environment is the world in which the agent operates in. At each time step, the agent is in a state, which refers to the representation of the environment perceived by the agent, and the agent takes an action, which is the decision taken by the agent based on the current state. The agent then receives a reward, which is the feedback signal from the environment that indicates the quality of the taken action. The policy represents the strategy for choosing actions at each state that the agent follows, and every policy has associated state value function, denoted as $V(s)$, and action value function, denoted as $Q(s,a)$. The state value function represents the expected cumulative reward an agent can obtain from being in a particular state following a policy, and the action value function represents the expected cumulative reward an agent can obtain by taking a specific action from a particular state following a policy.  

In the HVAC control problem, the goal is to minimize the energy consumption while maintaining the temperature in the comfort zone \cite{sierla2022review}. The state space consists of various environmental observations, such as indoor and outdoor temperatures. In this work, we do not consider direct control. Therefore, the action space is a list of different heating and cooling set points. In order for the agent to learn to achieve the goal, the reward is structured as two components: a penalization for indoor temperature being outside the comfort zone, and a penalization for consuming energy. Thus the reward is negative at all times.

\subsection{Q-Learning}

Q-Learning \cite{watkins1992q} is a well-established off-policy temporal-difference (TD) control algorithm in the field of reinforcement learning. In TD learning, the agent learns by bootstrapping from the current estimate of the value function. These methods sample from the environment and perform updates based on current estimates. TD learning employs the notion of TD error, which is the difference between the correct value $V_t^*$ and the current estimate value $V_t$, to make value function updates:

\begin{align}
    E_t &= {V_t}^* - V_t \\
        &=r_t + \gamma \cdot V_{t+1} - V_t
\end{align}

Off-policy learning means that the agent behavior policy during training is different from the learned policy. During training, Q-Learning employs $\epsilon$-greedy policy, a common strategy in reinforcement learning that balances exploitation and exploration for effective training. In the epsilon-greedy policy, the agent predominantly selects the action with the highest Q-value at each state. However, to promote exploration and prevent the agent from becoming stuck in sub-optimal actions, there is a small probability, $\epsilon$, of choosing a random action instead. The Q-value update rule is based on the Bellman equation, which states that the optimal Q-value for a state-action pair is equal to the immediate reward obtained plus the maximum expected future reward from the next state:

\begin{align}
    Q^{new}&(s_t, a_t) \leftarrow Q(s_t, a_t) + \\ 
        &\alpha(r_t + \gamma \cdot max_{a_{t+1}}Q(s_{t+1}, a_{t+1}) - Q(s_t, a_t)),
\end{align}
where $\alpha$ is the learning rate, and $\gamma$ is the discount factor that discounts future rewards. Please refer to Algorithm \ref{alg:q-learning} for the complete overview.

\begin{figure}[t!]
    \centering
    \includegraphics[width=\linewidth]{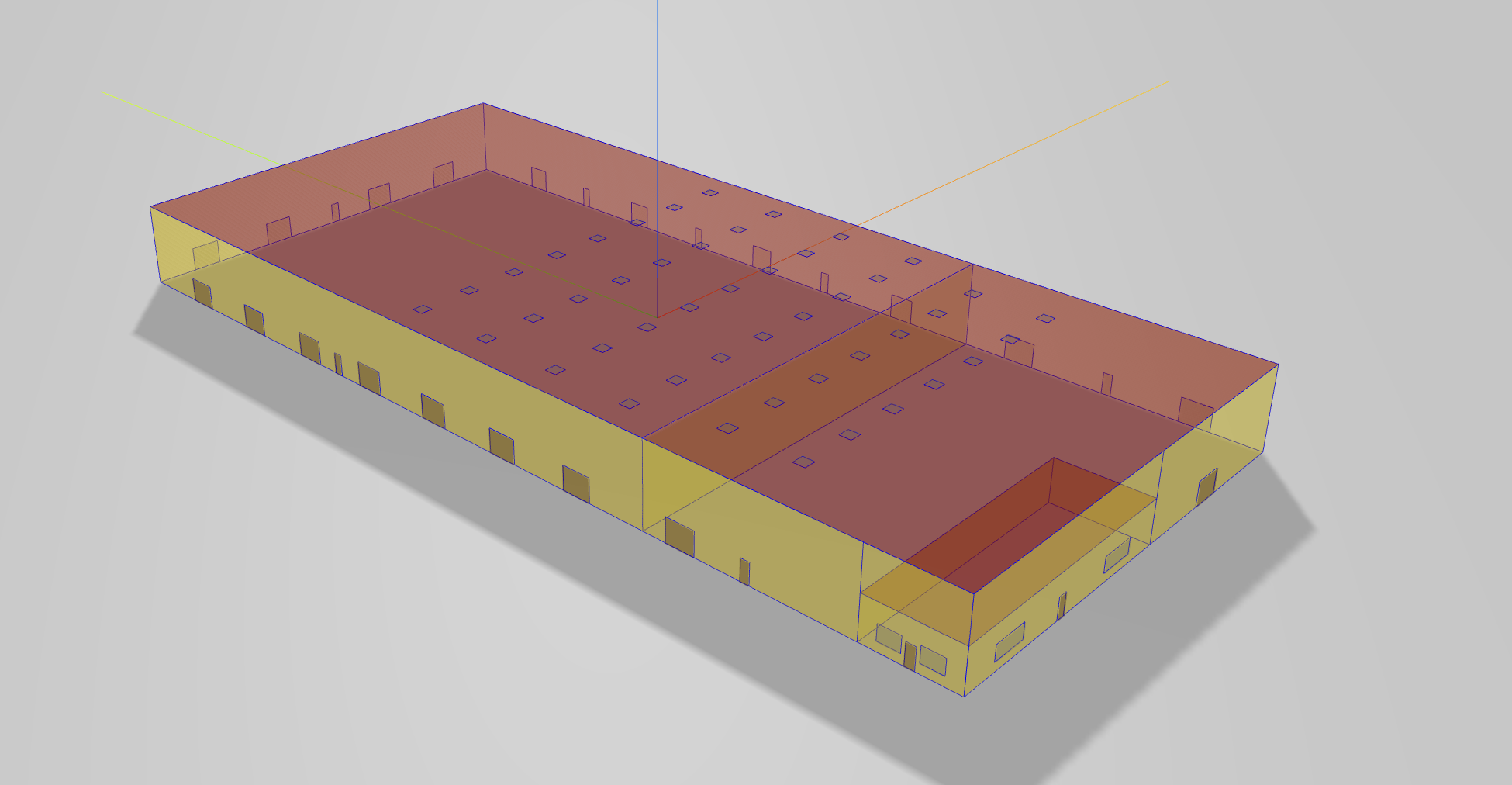}
    \caption{Warehouse Environment Visualization}
    \label{fig:warehouse}
\end{figure}

\begin{table}[t!]
\resizebox{\linewidth}{!}{%
\begin{tabular}{@{}lcc@{}}
\toprule
\multicolumn{1}{c}{\textbf{Description}} & \textbf{Notation} & \multicolumn{1}{l}{\textbf{Unit}} \\ \midrule
Site Outdoor Air Drybulb Temperature - Environment & $T_{out}$ & $^\circ C$ \\
Site Outdoor Air Relative Humidity - Environment & $H_{out}$ & $\%$ \\
Site Wind Speed - Environment & $V_{out}$ & $m/s$ \\
Site Wind Direction - Environment & $W_{out}$ & $\degree$ \\
Site Diffuse Solar Radiation Rate per Area - Environment & $S_{diffuse}$ & $W/m^2$ \\
Site Direct Solar Radiation Rate per Area - Environment & $S_{direct}$ & $W/m^2$ \\
Thermostat Heating Setpoint Temperature - Zone1 Office & $T_{office}^{hs}$ &  $^\circ C$\\
Thermostat Cooling Setpoint Temperature - Zone1 Office & $T_{office}^{cs}$ &  $^\circ C$\\
Zone Air Temperature - Zone1 Office & $T_{office}$ &  $^\circ C$\\
Air Relative Humidity - Zone1 Office & $H_{office}$ & $\%$ \\
Zone People Occupant Count  Zone1 Office & $C_{office}$ & $int$ \\
Thermostat Heating Setpoint Temperature - Zone2 Fine Storage & $T_{fs}^{hs}$ & $^\circ C$ \\
Thermostat Cooling Setpoint Temperature - Zone2 Fine Storage & $T_{fs}^{cs}$ & $^\circ C$ \\
Air Temperature - Zone2 Fine Storage & $T_{fs}$ &  $^\circ C$\\
Air Relative Humidity - Zone2 Fine Storage & $H_{fs}$ & $\%$\\
Thermostat Heating Setpoint Temperature - Zone3 Bulk Storage & $T_{bs}^{hs}$ & $^\circ C$ \\
Air Temperature - Zone3 Bulk Storage & $T_{bs}$ & $^\circ C$ \\
Air Relative Humidity - Zone3 Bulk Storage & $H_{bs}$ & $\%$ \\
Facility Total HVAC Electricity Demand Rate & $P_{total}$ & $W$ \\
\end{tabular}%
}
\caption{Warehouse Observation Space}
\label{tab:warehouse-observation}
\end{table}

\begin{table}[t!]
\resizebox{\linewidth}{!}{%
\begin{tabular}{ccc|cc|c}
\multicolumn{1}{l}{} & \multicolumn{2}{|c|}{\textbf{Office}} & \multicolumn{2}{c|}{\textbf{Fine Storage}} & \textbf{Bulk Storage} \\ \hline
\multicolumn{1}{c|}{\textbf{Action}} & $T_{office}^{hs}$ & $T_{office}^{cs}$ & $T_{fs}^{hs}$ & $T_{fs}^{cs}$ & $T_{bs}^{hs}$ \\ \hline
\multicolumn{1}{c|}{0} & 15 & 30 & 15 & 30 & 15 \\
\multicolumn{1}{c|}{1} & 16 & 29 & 16 & 29 & 16 \\
\multicolumn{1}{c|}{2} & 17 & 28 & 17 & 28 & 17 \\
\multicolumn{1}{c|}{3} & 18 & 27 & 18 & 27 & 18 \\
\multicolumn{1}{c|}{4} & 19 & 26 & 19 & 26 & 19 \\
\multicolumn{1}{c|}{5} & 20 & 25 & 20 & 25 & 20 \\
\multicolumn{1}{c|}{6} & 21 & 24 & 21 & 24 & 21 \\
\multicolumn{1}{c|}{7} & 22 & 23 & 22 & 23 & 22 \\
\multicolumn{1}{c|}{8} & 22 & 22 & 22 & 22 & 23 \\
\multicolumn{1}{c|}{9} & 21 & 21 & 21 & 21 & 24
\end{tabular}%
}
\caption{Warehouse Action Space}
\label{tab:warehouse-actions}
\end{table}

\subsection{Deep Q-Network (DQN)}

The Deep Q-Network (DQN) \cite{mnih2013playing} is a deep reinforcement learning algorithm developed by DeepMind for learning to play Atari 2600 games. It is a variant of Q-Learning, but instead of estimating Q-values and storing them in a Q-table, DQN employs neural networks to approximate the Q-functions. This enables the algorithm to handle continuous variables as observations. One challenge of using neural networks in reinforcement learning is the strong correlation between samples within the same trajectory, resulting in non-independent and non-identically distributed samples. DQN addresses this issue by introducing a technique called experience replay. Instead of directly feeding each sample from the agent's trajectory into the Q-function estimator network, the samples are stored in a replay buffer. The estimator then takes inputs from randomly sampled data points in the replay buffer, promoting independence and identical distribution of the input data. 
Please refer to Algorithm \ref{alg:dqn} for the complete overview.

\section{Experimental Setup} \label{sec:experiments}

We evaluate two simulated HVAC environments, a warehouse and a datacenter in multiple climates. We utilize the Sinergym \cite{2021sinergym} wrapper for the EnergyPlus \cite{crawley2001energyplus} simulation engine as the programmatic interface and simulator respectively. EnergyPlus requires an EnergyPlus Input File (IDF) to define the building and HVAC system structure and configuration and an EnergyPlus Weather File (EPW) which defines outdoor weather behavior and indirectly the simulation episode length.

\subsection{Warehouse}

The ``Warehouse'' environment (visualized in Figure \ref{fig:warehouse}) is a $4598m^2$ single-story commercial building that was developed by \cite{deru2011us} as a commercial reference building. The building is split into 3 HVAC control zones; the Office zone, the Bulk Storage zone, and the Fine Storage zone. The Warehouse observation space, $\mathbf{s}_{warehouse} = \{ \mathbf{s}_{out}, \mathbf{s}_{office}, \mathbf{s}_{bs}, \mathbf{s}_{fs} \}$, consists of outdoor temperature, humidity, wind, and solar environmental features, $\mathbf{s}_{out}$ and a set of features from each of the 3 zones; $\mathbf{s}_{office}$, $\mathbf{s}_{bs}$ and  $\mathbf{s}_{fs}$. The zone features include interior air temperatures, humidity, and heating and cooling setpoints. Please refer to Table \ref{tab:warehouse-observation} for the complete listing of Warehouse observation space features. The Warehouse action space, $\mathbf{a}_{warehouse}$ consists of a set of 10 integer indexed actions that correspond to heating and cooling setpoints in all 3 zones. The actions range from a relaxed policy of $\left( T_{office}^{hs}, T_{office}^{cs}, T_{fs}^{hs}, T_{fs}^{cs}, T_{fs}^{hs} \right) = \left( 15, 30, 15, 30, 15 \right)$ which minimizes energy consumption to an aggressive policy of  $\left( T_{office}^{hs}, T_{office}^{cs}, T_{fs}^{hs}, T_{fs}^{cs}, T_{fs}^{hs} \right) = \left( 21, 21, 21, 21, 24 \right)$ which maximizes thermal comfort. Please refer to Table \ref{tab:warehouse-actions} for the complete set of available actions.

\begin{figure}[t!]
    \centering
    \includegraphics[width=\linewidth]{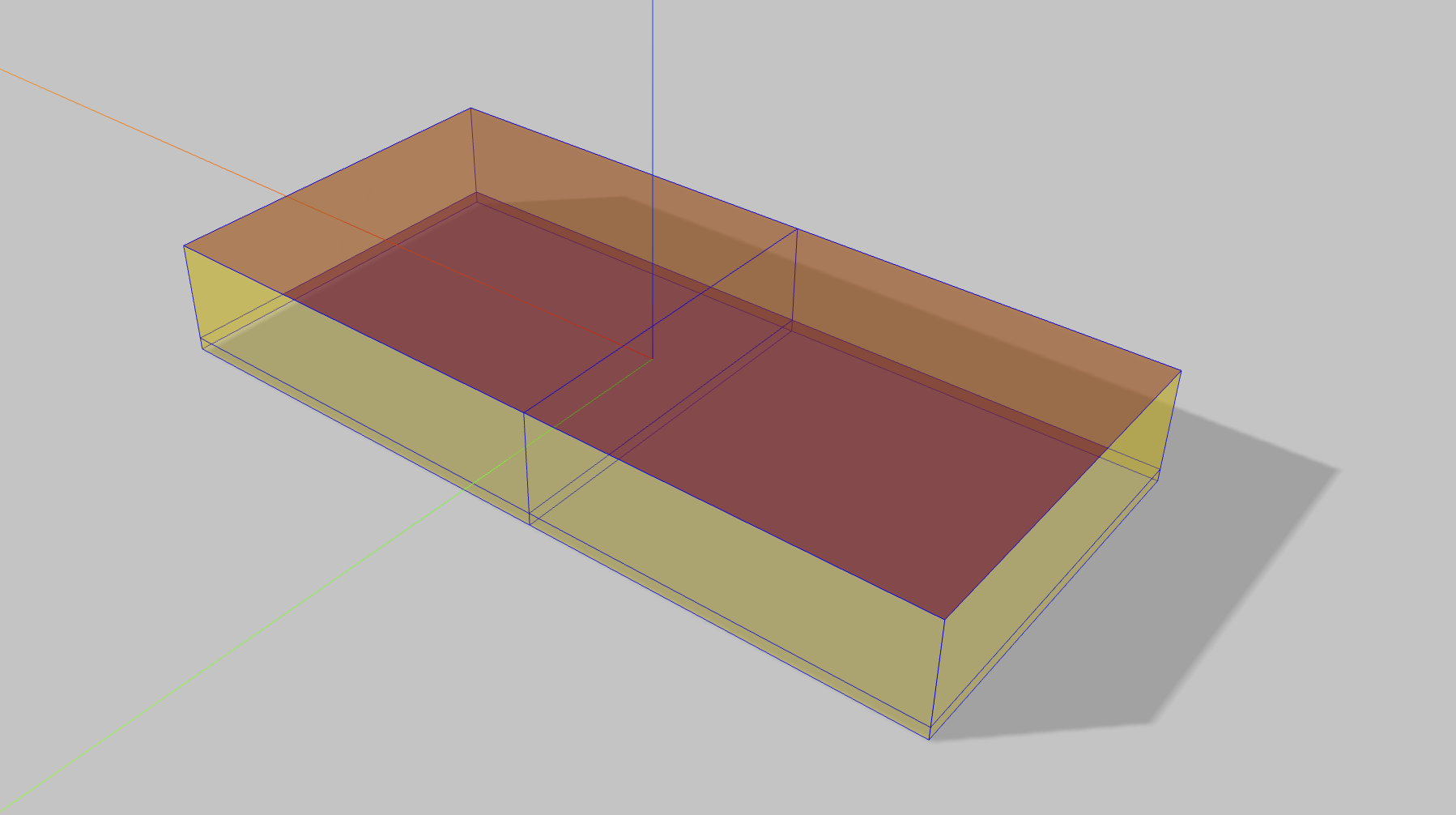}
    \caption{Datacenter Environment Visualization}
    \label{fig:datacenter}
\end{figure}

\begin{table}[t!]
\resizebox{\linewidth}{!}{%
\begin{tabular}{@{}lcc@{}}
\toprule
\multicolumn{1}{c}{\textbf{Description}} & \textbf{Notation} & \multicolumn{1}{l}{\textbf{Unit}} \\ \midrule
Site Outdoor Air Drybulb Temperature Environment & $T_{out}$ & $^\circ C$ \\
Site Outdoor Air Relative Humidity - Environment & $H_{out}$ & $\%$ \\
Site Wind Speed - Environment & $V_{out}$ & $m/s$ \\
Site Wind Direction - Environment & $W_{out}$ & $\degree$ \\
Site Diffuse Solar Radiation Rate per Area - Environment & $S_{diffuse}$ & $W/m^2$ \\
Site Direct Solar Radiation Rate per Area - Environment & $S_{direct}$ & $W/m^2$ \\
Thermostat Heating Setpoint Temperature - West Zone & $T_{wz}^{hs}$ & $^\circ C$ \\
Thermostat Cooling Setpoint Temperature West Zone & $T_{wz}^{cs}$ & $^\circ C$ \\
Air Temperature - West Zone & $T_{wz}$ & $^\circ C$ \\
Thermal Comfort Mean Radiant Temperature - West Zone & $T_{wz}^{cmr}$ & $^\circ C$ \\
Air Relative Humidity - West Zone & $H_{wz}$ & \% \\
Thermal Comfort Clothing Value - West Zone & $T_{wz}^{ccv}$ & $^\circ C$ \\
Thermal Comfort Fanger Model PPD - West Zone & $T_{wz}^{cfm}$ & \% \\
People Occupant Count - West Zone & $C_{wz}$ & $int$ \\
People Air Temperature - West Zone & $T_{wz}^{pa}$ & $^\circ C$ \\
Thermostat Heating Setpoint Temperature - East Zone & $T_{ez}^{hs}$ & $^\circ C$ \\
Thermostat Cooling Setpoint Temperature - East Zone & $T_{ez}^{cs}$ & $^\circ C$ \\
Air Temperature - East Zone & $T_{ez}$ & $^\circ C$ \\
Thermal Comfort Mean Radiant Temperature - East Zone & $T_{ez}^{cmr}$ & $^\circ C$ \\
Air Relative Humidity-  East Zone & $T_{ez}$ &  \% \\
Thermal Comfort Clothing Value - East Zone & $T_{ez}^{ccv}$ & $^\circ C$ \\
Thermal Comfort Fanger Model PPD - East Zone & $T_{ez}^{cfm}$ & \% \\
People Occupant Count- East Zone & $C_{ez}$ & $int$ \\
People Air Temperature - East Zone & $T_{ez}^{pa}$ & $^\circ C$ \\
Facility Total HVAC Electricity Demand Rate & $P_{total}$ & $W$ \\
\end{tabular}%
}
\caption{Datacenter Observation Space}
\label{tab:datacenter-observation}
\end{table}

\begin{table}[t!]
\centering
\resizebox{0.7\linewidth}{!}{%
\begin{tabular}{ccc|cc}
\multicolumn{1}{l}{} & \multicolumn{2}{|c|}{\textbf{West Zone}} & \multicolumn{2}{c}{\textbf{East Zone}} \\ \hline
\multicolumn{1}{c|}{\textbf{Action}} & $T_{wz}^{hs}$ & $T_{wz}^{cs}$ & $T_{ez}^{hs}$ & $T_{ez}^{cs}$ \\ \hline
\multicolumn{1}{c|}{0} & 15 & 30 & 15 & 30 \\
\multicolumn{1}{c|}{1} & 16 & 29 & 16 & 29 \\
\multicolumn{1}{c|}{2} & 17 & 28 & 17 & 28 \\
\multicolumn{1}{c|}{3} & 18 & 27 & 18 & 27 \\
\multicolumn{1}{c|}{4} & 19 & 26 & 19 & 26 \\
\multicolumn{1}{c|}{5} & 20 & 25 & 20 & 25 \\
\multicolumn{1}{c|}{6} & 21 & 24 & 21 & 24 \\
\multicolumn{1}{c|}{7} & 22 & 23 & 22 & 23 \\
\multicolumn{1}{c|}{8} & 22 & 22 & 22 & 22 \\
\multicolumn{1}{c|}{9} & 21 & 21 & 21 & 21 
\end{tabular}%
}
\caption{Datacenter Action Space}
\label{tab:datacenter-actions}
\end{table}

\subsection{Datacenter}

The ``Datacenter'' environment (visualized in Figure \ref{fig:datacenter}) is a $491.3 m^2$ building divided into two zones; the West and East zones. The observation and action spaces are constructed similarly to the Warehouse environment. The observation space, $\mathbf{s}_{datacenter} = \{ \mathbf{s}_{out}, \mathbf{s}_{wz}, \mathbf{s}_{ez}\}$, consists of outdoor temperature, humidity, wind, and solar environmental features alongside setpoint and zone-specific features $\mathbf{s}_{wz}$ and $\mathbf{s}_{ez}$. Please refer to Table \ref{tab:datacenter-observation} for the complete listing of Datacenter observation variables. The action space follows the same structure as the Warehouse environment, moving from a relaxed policy to an aggressive policy. The complete action space is provided in Table \ref{tab:datacenter-actions}.

\subsection{Locations}

We investigate each building in two different geographic locations and climates within the United States. The first location, ``Hot'', is a hot desert (Köppen BWh) climate, simulated using weather data collected from the Davis-Monthan Air Force Base outside of Tuscon, Arizona. The second location, ``Cool'', is a Mediterranean climate (Köppen Csb), simulated using weather data collected from the William R. Fairchild International Airport in Port Angeles, Washington. 

\subsection{Evaluation}

To investigate the algorithmic performance of HVAC control we first construct a set of "training" and "evaluation" episodes following the D4RL evaluation protocol \cite{fu2021d4rl}. In each location, we construct an 80/20\% split of the weather data to be used for training and evaluation respectively. Additionally, we use the power consumption and temperature violation evaluation metrics proposed by \cite{wei2017deep}. The power consumption metric simply measures the total amount of energy consumed in the building during evaluation. The temperature violation metric measures the total time that the system spent outside of the target temperature comfort zone defined by minimum and maximum target temperatures,  $T_{c}^{min}$ and $T_{c}^{max}$. In all environments, $T_{c}^{min}$ was set as $18^\circ C$ and $T_{c}^{max}$ as $27^\circ C$.

\subsection{Model Implementation \& Hyperparameters}

The Q-Learning model uses a learning rate of 0.0001 and an initial epsilon value of 1, with an annealing rate of 0.12 and a final value of 0.1. For DQN we utilized the StableBaselines implementation with default hyperparameters. Both algorithms utilized the same reward function provided by Sinergym \cite{2021sinergym},

\begin{equation}
    r_t=-\omega \lambda_P P_t-(1-\omega) \lambda_T\left(\left|T_t-T_{c}^{max}\right|+\left|T_t-T_{c}^{min}\right|\right),
\end{equation}
which is a linear combination of a negative energy consumption term and a negative temperature violation term. $P_t$ is the current energy consumption, $T_t$ is the current temperature, $T_{c}^{min}$ and $T_{c}^{max}$ represents the temperature comfort range, and temperature violation penalty is 0 if the temperature is within this range \cite{2021sinergym}. $\omega$ is the relative weighting parameter, it is set to 0.5 by default so that both penalty terms have equal weights, and $\lambda_P$ and $\lambda_T$ are scaling parameters to establish proportional concordance between energy and temperature \cite{2021sinergym}.

\begin{figure}[t!]
    \centering
    \includegraphics[width=0.85\linewidth]{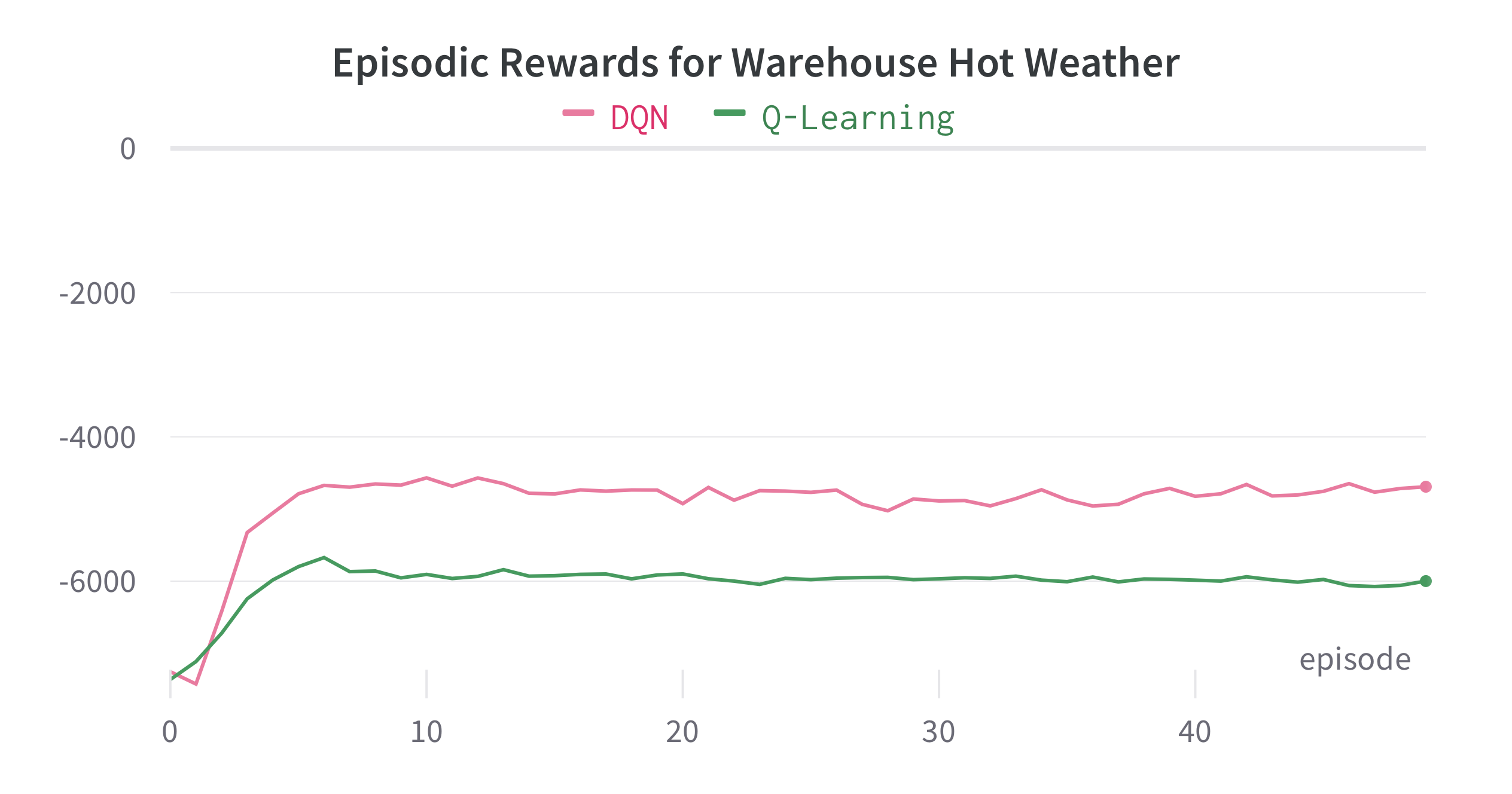}
    \caption{Warehouse Hot Weather Training Curve}
    \label{fig:warehouse-hot}
\end{figure}

\begin{figure}[t!]
    \centering
    \includegraphics[width=0.85\linewidth]{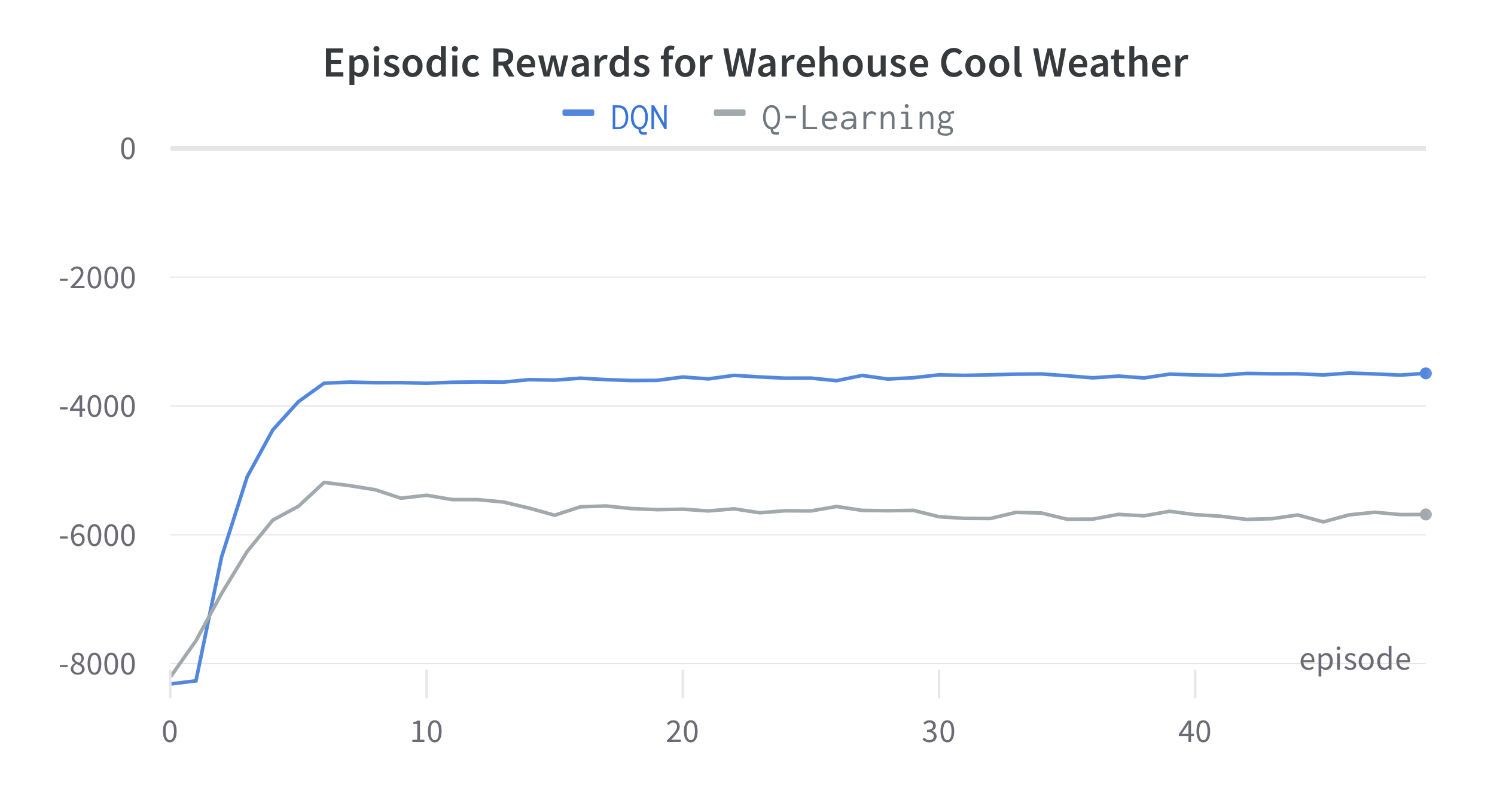}
    \caption{Warehouse Cool Weather Training Curve}
    \label{fig:warehouse-cool}
\end{figure}


\begin{table}[]
\centering
\resizebox{\linewidth}{!}{%
\begin{tabular}{@{}c|cc@{}}
\toprule
 & \multicolumn{2}{c}{\textbf{Hot Weather/Datacenter}} \\ \midrule
\textbf{Method} & \textbf{Energy (kWh)} & \textbf{Temp. Violation (\%)} \\ \midrule
Fixed Agent & $154233$ & $\mathbf{0.00}$ \\
Q-Learning & $\mathbf{143545}$ & $26.50$ \\
DQN & $149544$ & $6.96$ \\ \bottomrule
\end{tabular}%
}
\caption{Datacenter Environment Hot Weather Results}
\label{tab:datacenter-hot}
\end{table}

\begin{table}[]
\resizebox{\linewidth}{!}{%
\begin{tabular}{@{}c|cc@{}}
\toprule
\textbf{} & \multicolumn{2}{c}{\textbf{Hot Weather/Warehouse}} \\ \midrule
\textbf{Method} & \textbf{Energy (kWh)} & \textbf{Temp. Violation (\%)} \\ \midrule
Fixed Agent & $28429$ & $\mathbf{4.64}$ \\
Q-Learning & $\mathbf{24621}$ & $36.19$ \\
DQN & $25639$ & $10.85$ \\ \bottomrule
\end{tabular}%
} 
\caption{Warehouse Environment Hot Weather Results}
\label{tab:warehouse-hot}
\end{table}

\begin{table}[]
\resizebox{\linewidth}{!}{%
\begin{tabular}{@{}c|cc@{}}
\toprule
\textbf{} & \multicolumn{2}{c}{\textbf{Cold Weather/Warehouse}} \\ \midrule
\textbf{Method} & \textbf{Energy (kWh)} & \textbf{Temp. Violation (\%)} \\ \midrule
Fixed Agent & $26558$ & $\mathbf{6.95}$ \\
Q-Learning & $\mathbf{24537}$ & $34.63$ \\
DQN & $25317$ & $13.31$ \\ \bottomrule
\end{tabular}%
} 
\caption{Warehouse Environment Cold Weather Results}
\label{tab:warehouse-cold}
\end{table}


\section{Results} \label{sec:results}
We selected a fixed-action agent as the baseline. This agent consistently chooses actions with heating and cooling setpoints within the comfort zone. We compare both the Q-Learning agent and the DQN agent against this baseline in the warehouse environment and the datacenter environment with hot weather. 

Most of the environment variables are continuous variables, and Q-Learning cannot handle continuous state variables by nature. To address this issue, we performed tile coding on the continuous variables when training the Q-Learning agent, which is a technique that discretizes continuous state space into uniform bands of values as tiles. Another issue with Q-Learning is that if the state space dimensionality becomes too high, then the Q table would exceed the memory limit. We only kept the core observation variables such as temperature and humidity to prevent this from happening. The tile resolution is roughly 5{\degree}C for temperature tiles, and humidity tiles are on a similar scale. As for DQN, it can take continuous observations, and it uses a neural network to approximate the Q function instead of storing the Q values in a table, so no additional processing is required. Both agents were trained over 50 episodes and tested on unseen weather data.

In both the datacenter and the warehouse environments, as shown in Table \ref{tab:datacenter-hot}, Table \ref{tab:warehouse-hot} and Table \ref{tab:warehouse-cold}, we observed that DQN displayed the best overall performance. The baseline agent guarantees maximum comfort, so it always has the lowest percentage of temperature violations, but at the same time, maximum comfort leads to the highest power usage. In terms of energy consumption, the Q-Learning agent consumes the least energy, yet the temperature violation percentages are very high compared to the other two. DQN shows a moderate temperature violation percentage while saving power at the same time. In addition, we also conducted experiments in cool weather for the warehouse environment, and the performance pattern amongst these agents remains the same. Please refer to Table \ref{tab:warehouse-hot}, Table \ref{tab:warehouse-cold}, Figure \ref{fig:warehouse-hot}, and Figure \ref{fig:warehouse-cool} for details. We further performed ablation studies in the warehouse environment to understand the impact of state space choices, reward weightings, and tile coding density. 

\subsection{Observation Space Ablation}

We categorized the available observations into four groups: environment variables (related to temperature and humidity), energy variables (tracking energy consumption), action variables (reflecting HVAC setpoint changes), and auxiliary variables (such as people occupancy). We conducted ablation experiments by gradually including additional categories of variables in the training process, starting with only environment variables and then introducing subsequent categories in the order mentioned.

Due to memory limitations, we limited the Q-Learning agent experiments to using only environment variables and environment variables combined with energy variables. As shown in Table \ref{tab:ql-state-ablation} and Table \ref{tab:dqn-state-ablation}, across the six sets of experiments conducted with both Q-Learning and DQN, we observed an interesting pattern. While the power consumption tended to decrease gradually as more variables were included in the training, the percentage of temperature violations consistently increased, resulting in poorer overall performance.

This pattern can be attributed to the dimensionality of the state space. In the warehouse environment, there are a total of 19 observation variables. The complexity of both the Q-table used in Q-Learning and the neural network employed in DQN is insufficient to appropriately capture the dynamics among these variables. Consequently, restricting the observations to only the core environment variables yielded stronger performance. 

\subsection{Reward Weighting Ablation}

 The experiments aimed to investigate the effects of modifying the reward term weighting. Specifically, the energy penalty weight was increased to 0.75 and decreased to 0.25. The DQN agent was used for these experiments due to its superior performance compared to Q-Learning. The detailed results are shown in Table \ref{tab:dqn-reward-weights}.

The results of the experiments demonstrated that when the energy consumption reward weight was increased, the agent's choices would consume less energy, but at the cost of the temperature frequently exceeding the comfort zone. Conversely, when the weight was decreased, the agent's choice consumed more energy but managed to keep the temperature within the comfort zone more consistently.

These findings indicate a trade-off between energy conservation and temperature violation, and the reward weights should be adjusted according to the use case. For instance, in an environment that is more strict on temperature violations, such as datacenters, it would be better to decrease the weight on energy consumption.

\subsection{Tile Coding Density Ablation}

Higher density in tile coding means a more fine-grained approximation of the original continuous state space, which could potentially lead to better performance. Limited by memory size, we conduct an additional set of experiments exploring different tile coding density configurations with only environment observation variables. We increased the tile coding density to roughly 2{\degree}C per tile for temperature variables and updated the humidity variables accordingly. The results in Table \ref{tab:ql-discretization-density} demonstrate lower power consumption with increased tile coding density, but this reduction incurs significantly more temperature violations.  

\begin{table}[]
\resizebox{\linewidth}{!}{%
\begin{tabular}{@{}cccc|cc@{}}
\toprule
\multicolumn{4}{c|}{\textbf{Observation Variables}} & \multicolumn{2}{c}{\textbf{Q-Learning}} \\ \midrule
\textbf{Env} & \textbf{Energy} & \textbf{Action} & \textbf{Aux} & \textbf{Energy (kWh)} & \textbf{Temp.Violation (\%)} \\ \midrule
\checkmark & - & - & - & $\mathbf{24621}$ & $\mathbf{36.19}$ \\
\checkmark & \checkmark & - & - & $24644$ & $39.64$ \\ \bottomrule
\end{tabular}%
}
\caption{Q-Learning Observation Space Ablation}
\label{tab:ql-state-ablation}
\end{table}

\begin{table}[]
\resizebox{\linewidth}{!}{%
\begin{tabular}{@{}cccc|cc@{}}
\toprule
\multicolumn{4}{c|}{\textbf{Observation Variables}} & \multicolumn{2}{c}{\textbf{DQN}} \\ \midrule
\textbf{Env} & \textbf{Energy} & \textbf{Action} & \textbf{Aux} & \textbf{Energy (kWh)} & \textbf{Temp.Violation (\%)} \\ \midrule
\checkmark & - & - & - & $26906$ & $\mathbf{6.07}$ \\
\checkmark & \checkmark & - & - & $26080$ & $8.62$ \\
\checkmark & \checkmark & \checkmark & - & $25879$ & $11.00$ \\
\checkmark & \checkmark & \checkmark & \checkmark & $\mathbf{25639}$ & $10.85$ \\ \bottomrule
\end{tabular}%
}
\caption{DQN Observation Space Ablation}
\label{tab:dqn-state-ablation}
\end{table}

\begin{table}[]
\resizebox{\linewidth}{!}{%
\begin{tabular}{@{}ccc@{}}
\toprule
\multicolumn{1}{c|}{\textbf{Reward Weight}} & \textbf{Energy (kWh)} & \textbf{Temp. Violation (\%)} \\ \midrule
\multicolumn{1}{c|}{0.75} & $\mathbf{25447}$ & $15.89$ \\
\multicolumn{1}{c|}{0.5} & $25639$ & $10.85$ \\
\multicolumn{1}{c|}{0.25} & $26222$ & $\mathbf{8.99}$ \\ \bottomrule
\end{tabular}%
}
\caption{DQN Reward Ablation}
\label{tab:dqn-reward-weights}
\end{table}

\begin{table}[]
\resizebox{\linewidth}{!}{%
\begin{tabular}{@{}c|cc@{}}
\toprule
\textbf{Tile Coding Density ($\degree C$/tile)}  & \textbf{Energy (kWh)} & \textbf{Temp. Violation (\%)} \\ \midrule
5 & $24621$ & $\mathbf{36.19}$ \\
2 & $\mathbf{23221}$ & $52.00$ \\ \bottomrule
\end{tabular}%
}
\caption{Q-Learning Tile Coding Density Ablation}
\label{tab:ql-discretization-density}
\end{table}

\section{Conclusion}

This work provides practical insights into training RL agents to control HVAC systems. The results demonstrate the trade-off between energy conservation and temperature violation and explore the influence of state space choices when operating with high-dimensional observation spaces. When using simple function approximators, selecting only the core observation components will likely lead to better performance. We also demonstrated DQN to be a good choice for HVAC control tasks due to its balanced performance. Overall, the deployment of RL agents for HVAC control offers promising prospects for substantial environmental benefits and cost savings in real-world settings. 

\clearpage

\bibliographystyle{IEEEtran}
\bibliography{cites.bib} 

\end{document}